\pdfoutput=1

\documentclass[11pt]{article}

\usepackage[]{acl}

\usepackage{times}
\usepackage{latexsym}

\usepackage[T1]{fontenc}

\usepackage[utf8]{inputenc}

\usepackage{microtype}

\usepackage{inconsolata}

\usepackage{makecell}
\usepackage{multirow}
\usepackage{amsmath}
\usepackage{amssymb}
\usepackage{graphicx}
\usepackage{booktabs}
\usepackage{subcaption}
\usepackage{xcolor}

\newcommand{\red}[1]{\textcolor{red}{#1}}
\definecolor{darkgreen}{rgb}{0.0, 0.8, 0.0}
\newcommand{\green}[1]{\textcolor{darkgreen}{#1}}

    \title{Towards Reasoning in Large Language Models via  \\
    Multi-Agent Peer Review Collaboration}

\author{
Zhenran Xu\textsuperscript{1},
Senbao Shi\textsuperscript{1},
Baotian Hu\textsuperscript{1}\thanks{$\quad$Corresponding author.}~,
Jindi Yu\textsuperscript{1},
Dongfang Li\textsuperscript{1},
Min Zhang\textsuperscript{1},
Yuxiang Wu\textsuperscript{2} \\
\textsuperscript{1}Harbin Institute of Technology (Shenzhen) \quad \textsuperscript{2}University College London\\
\texttt{xuzhenran@stu.hit.edu.cn, shisenbaohit@gmail.com} \\
\texttt{\{hubaotian, zhangmin2021\}@hit.edu.cn} \\
\texttt{22S051013@stu.hit.edu.cn, crazyofapple@gmail.com, yuxiang.wu@cs.ucl.ac.uk} \\
}

\begin{document}
\maketitle
\begin{abstract}
Large Language Models (LLMs) have shown remarkable capabilities in general natural language processing tasks but often fall short in complex reasoning tasks. 
Recent studies have explored human-like problem-solving strategies, such as \textit{self-correct}, to push further the boundary of single-model reasoning ability. 
In this work, we let a single model ``step outside the box'' by engaging multiple models to correct each other.
We introduce a multi-agent collaboration strategy that emulates the academic peer review process. 
Each agent independently constructs its own solution, 
provides reviews on the solutions of others, and assigns confidence levels to its reviews.
Upon receiving peer reviews, agents revise their initial solutions.
Extensive experiments on three different types of reasoning tasks
show that our collaboration approach delivers superior accuracy across all ten datasets compared to existing methods.
Further study underscores the effectiveness of integrating confidence in reviews, demonstrates the superiority of feedback exchange over mere solution sharing, and highlights the role of capability and diversity in fostering successful collaboration.
Code is available at \url{https://github.com/HITsz-TMG/Multi-agent-peer-review}.



\end{abstract}

\begin{figure}[t]
    \centering
    \includegraphics[width=\linewidth]{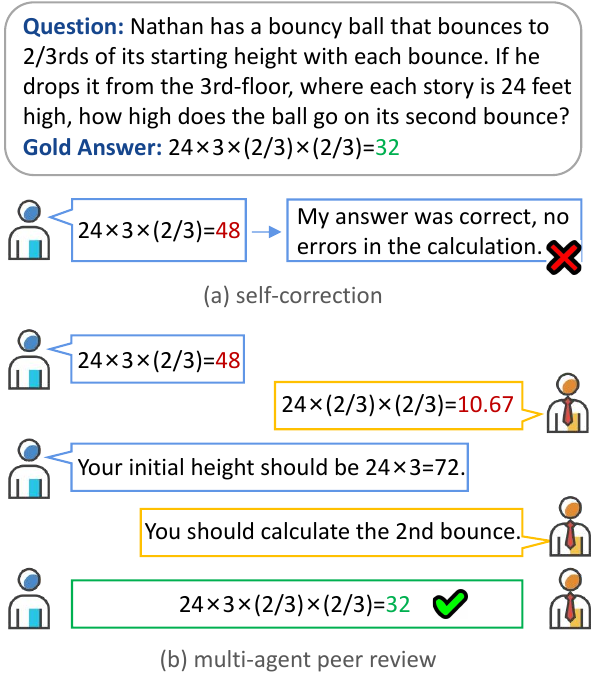}
    \caption{Comparison of self-correction~\cite{huang2023noselfcorrect} and our multi-agent collaboration method. While a single LLM struggles to develop new ideas and self-correct its response, our approach enables multiple agents to engage in mutual review and correction, leading to an accurate consensus.}
    \label{fig:intro}
\end{figure}

\section{Introduction}

\begin{quote}
\textit{``Lookers-on see the chess game better than players.''
-- Chinese Proverb
}
\end{quote}

Large language models (LLMs) have demonstrated remarkable capabilities in a wide range of natural language processing (NLP) tasks~\cite{touvron2023llama,openai2023gpt4,anil2023palm}.
The emergent abilities of LLMs have showcased promising results across a multitude of benchmarks~\cite{wei2022emergent}.
Among all rising capabilities, complex reasoning opens up opportunities for building a large spectrum of applications upon LLMs, effectively making LLMs the next-generation computation platform~\cite{fu2023chainofthought}.
Current studies have explored human-like problem-solving strategies to push further the boundary of \textit{single-model} reasoning ability, such as chain-of-thought prompting~\cite{cot_neurips} and self-consistency~\cite{wang2023selfconsistency}.

Amidst this background, inspired by the typical human learning strategy of ``self-reflection'',
the idea of ``self-correction'' has gained significant research interest, where an LLM automatically reflects on its response and refines based solely on its inherent capabilities~\cite{madaan2023selfrefine,shinn2023reflexion,pan2023automatically}.
However, \citet{liang2023encouraging} suggest that, once the LLM has formulated initial solutions, it cannot generate novel thoughts through self-reflection.
Considering the example in Figure~\ref{fig:intro},
the initial answer of the LLM is incorrect.
Despite conducting a self-review and attempting to identify problems,
the LLM still assesses its answer as being correct.
\citet{huang2023noselfcorrect} further note that a single LLM struggles to self-correct its response without external feedback,
which leads us to the motivation behind our work:
Why not let a single LLM ``jump outside the box'' and let multiple LLMs provide feedback to each other?

To this end, we propose a \textit{multi-agent} collaboration strategy that mirrors the academic peer review process.
For a given question, each agent first independently submits its own solution.
Then the agents perform a systematic review of their peers' solutions. 
They not only examine the reasoning process step by step but also attach a confidence score to quantify the reliability of their feedback.
Drawing on the insights from the peer review, each agent finally refines its initial submission and creates an improved version.  
In the example shown in Figure~\ref{fig:intro},
our collaboration strategy enables multiple agents to engage in mutual review and correction, leading to an accurate consensus.
This iterative cycle of creation, review, and revision is akin to the collaborative efforts seen in human learning and problem-solving.

We conduct extensive experiments across three types of reasoning tasks:
mathematical reasoning, commonsense reasoning, and symbolic reasoning.
The results illustrate that our collaboration method achieves superior performance across all ten datasets compared to previous strong single-agent and multi-agent baselines.
Further analysis underscores the effectiveness of expressing uncertainty in the reviews,
demonstrates the advantage of sharing feedback rather than solutions,
and suggests that models with minimal capability differences but high diversity can collaborate better.

The contributions of this work are threefold:
\begin{itemize}
    \item By letting a single agent ``think outside the box'', we propose a multi-agent collaboration method that emulates the peer review process -- ``creation, review, and revision''.
    \item Our collaboration method consistently outperforms previous strong baselines across three reasoning tasks and ten datasets.
    \item We demonstrate the efficacy of integrating confidence and sharing feedback, suggesting a viable path forward the optimization of LLM-based multi-agent systems.
\end{itemize}

\begin{figure*}[t]
    \centering
    \includegraphics[width=\textwidth]{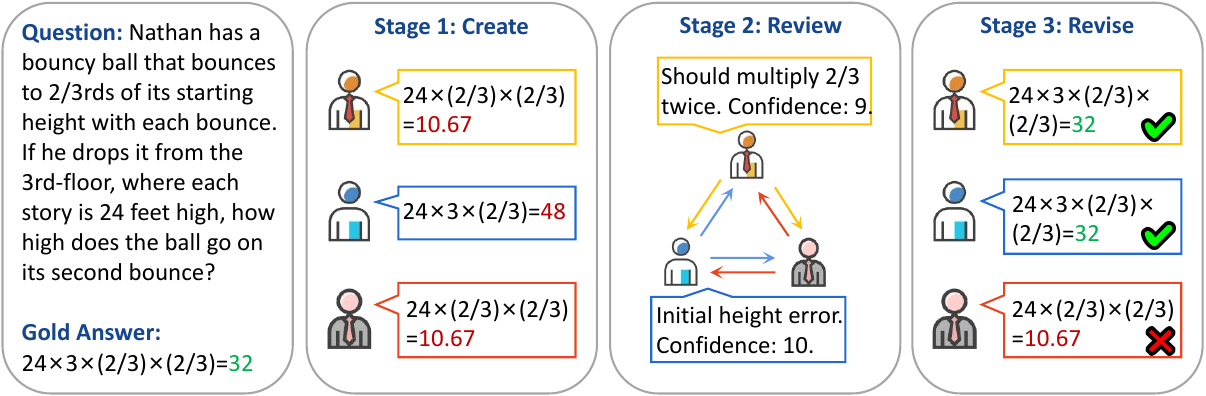}
    \caption{Illustration of our multi-agent peer review collaboration with an example in GSM8K~\cite{cobbe2021gsm}. The process consists of three stages -- ``creation, review, and revision''. The process starts with three incorrect initial responses. Through a series of human-like discussions and critiques among the agents,
    two of them arrive at the same correct solution.}
    \label{fig:method}
\end{figure*}

\section{Related Work}

With language models scaling up, the emergent abilities have showcased remarkable results across a variety of domains and tasks, challenging our understanding of learning and cognition~\cite{bubeck2023sparks,openai2023gpt4}.
Among these rising capabilities, 
\textbf{complex reasoning} draws significant interest for its central role in building autonomous language agents and the next-generation computation platform~\cite{xi2023rise,fu2023chainofthought}.

To push further the boundary of \textbf{single-agent reasoning},
researchers have adopted strategies akin to teaching kids to reason.
For example,
the chain-of-thought prompting~\cite{cot_neurips} employs rationales to steer LLMs through a multi-step reasoning process;
the self-consistency method~\cite{wang2023selfconsistency} selects the most likely answer from multiple reasoning paths by majority voting;
the plan-and-solve prompting~\cite{wang-etal-2023-plan} and the least-to-most prompting~\cite{zhou2023leasttomost} both apply a divide-and-conquer strategy: break down a complex problem into a series of simpler sub-problems and then solve them in sequence.

Amidst this backdrop, 
a prevailing strategy to address reasoning errors is \textbf{self-correction},
mirroring a typical human learning strategy of self-reflection~\cite{pan2023automatically}.
This approach involves an LLM attempting to refine its prior response based solely on its inherent capabilities, without the crutch of external feedback.
Contrary to the optimism about self-correction~\cite{madaan2023selfrefine,shinn2023reflexion,kim2023language},
\citet{huang2023noselfcorrect} find that an LLM struggles to correct its response without external feedback.
\citet{liang2023encouraging} also conclude that, once the LLM has formulated initial solutions, it cannot generate novel thoughts through self-reflection.
In this work,
we let a single LLM ``jump outside the box'' and change its viewpoint from a participant to a critical observer by enabling multiple LLMs to correct each other.

\textbf{Multi-agent collaboration} for task solving is still in its preliminary stages, presenting vast opportunities for exploration.
This approach aligns with the Society of Mind (SoM) theory~\cite{minsky1988society}, which suggests that intelligence arises from the interaction of computational modules, achieving collective goals beyond the capabilities of individual modules.
CAMEL~\cite{li2023camel} proposes a cooperative agent framework known as role-playing, enabling agents to cooperate on complex tasks.
Such role-playing collaboration has been extended to software development~\cite{qian2023chatdev} and text evaluation~\cite{chan2023chateval,li2023prd}.
\citet{liang2023encouraging} and \citet{du2023improving} have delved into multi-agent debates for translation and arithmetic problems.
Corex~\cite{sun2023corex} has explored three collaboration paradigms in reasoning -- debate, code review and retrieve.
Through these multi-agent simulations,
\citet{zhang2023exploring} observe human-like social behaviors typical of human collaboration.
To this end,
we propose a human-mimicking multi-agent collaboration process,
wherein multiple agents engage in peer discussions and provide feedback to one another.

\begin{table*}[ht]
\centering
\begin{tabular}{llccl}
\toprule
\textbf{Dataset}     & \textbf{Domain} & \textbf{\# Samples} & \textbf{Avg. words} & \textbf{Answer} \\ \midrule 
GSM8K~\cite{cobbe2021gsm}                & Math            & 1319                & 46.9                & Number          \\  
SVAMP~\cite{patel-etal-2021-svamp}                & Math            & 1000                & 31.8                & Number          \\  
AQuA~\cite{ling-etal-2017-aqua}                 & Math            & 254                 & 51.9                & Option          \\  
MultiArith~\cite{roy-roth-2015-multiarith}           & Math            & 600                 & 31.8                & Number          \\  
AddSub~\cite{hosseini-etal-2014-addsub}               & Math            & 395                 & 31.5                & Number          \\  
SingleEq~\cite{koncel-kedziorski-etal-2015-singleeq}             & Math            & 508                 & 27.4                & Number          \\  
ARC-c~\cite{clark2018arc}                & CS              & 1172                & 22.3                & Option          \\  
StrategyQA~\cite{geva-etal-2021-strategyqa}           & CS              & 2290                & 9.6                 & Yes / No        \\  
Colored Objects~\cite{suzgun-etal-2023-bbh}    & Sym.            & 250                 & 37.2                & Option          \\  
Penguins~\cite{suzgun-etal-2023-bbh}             & Sym.            & 146                 & 86.7                & Option          \\  
\bottomrule
\end{tabular}
\caption{Details of ten evaluation datasets. Math: mathematical reasoning. CS: commonsense reasoning. Sym.: symbolic reasoning.}
\label{tab:statistics}
\end{table*}

\section{Methodology}
\label{sec:method}

\textbf{Overview.} In this work, we introduce a multi-agent collaboration strategy that mirrors the academic peer review process. 
The overall process is illustrated in Figure~\ref{fig:method}.
The process consists of three stages -- ``creation, review, and revision''.
In stage 1, for a given question, each agent first independently submits its own solution.
In stage 2, each agent performs a systematic review of its peers' solutions. 
In stage 3, upon receiving the reviews, each agent refines its initial solution and re-submits the improved version. 
An example of these stages is provided in Section~\ref{sec:cases}.
We next elaborate these stages in detail.

\textbf{Stage 1: Create.} 
The peer review process starts with each agent submitting its own solution.
Specifically, when presented with a question $q$, 
A group of $n$ agents, denoted as ${A_1,A_2,...,A_n}$, perform chain-of-thought reasoning.
Each agent generates its own reasoning chains and corresponding answers $a_i$.
Take Figure~\ref{fig:method} as an example,
the initial answers of three agents are all incorrect.

\textbf{Stage 2: Review.}
We next feed each agent $A_i$ with the solution of its peers (i.e., $A_j$, where $j \neq i$), one at a time, to write reviews $r_{ij}$.
Note that $r_{ij}$ denotes the review \textit{from} agent $A_i$ \textit{to} agent $A_j$.
In current review forms,
reviewers do more than just provide overall assessments; they also indicate their certainty about their evaluations.
Mirroring such real-life review practices,
the agents not only examine the reasoning process step by step but also attach a confidence score to quantify the reliability of their feedback.
To achieve this, we concatenate each peer solution with the following instruction:
\textit{Please examine this agent's reasoning process step by step and offer feedback on its reasoning. You can rate your confidence in your feedback on a scale from 1-10, where 10 indicates the highest level of confidence.}

In the example shown in Figure~\ref{fig:method},
one agent accurately determines the initial height but forgets to multiply the ratio twice. Conversely, another agent multiplies the ratio twice but gets the initial height wrong. 
Although their initial answers are wrong,
they can provide effective feedback confidently to each other, based on the difference between their solutions and those of their peers.

\textbf{Stage 3: Revise.}
We finally feed each agent $A_i$ with the reviews from its peers (i.e., $r_{ji}$), all at once, to refine the initial solutions.
The agent considers both the peer solutions in dialogue history and the received peer reviews.
To achieve this, we concatenate the reviews with the following instruction:
\textit{Using other agents' solutions and feedbacks as additional information, can you provide your answer to the problem? }
In the example illustrated in Figure~\ref{fig:method},
out of three agents, two successfully integrate accurate feedback into their final responses. This results in both agents converging on the same correct solution after revision. 

Unlike prior studies such as~\citet{du2023improving} and \citet{sun2023corex}, we do not explicitly ask the agents to reach a consensus at the end of the review. 
The final prediction is determined through a majority vote among the $n$ participating agents.
This approach ensures the impartiality and balance of our peer review process.

\section{Experiment}

\begin{table*}[t]
\centering
\begin{tabular}{l c c c c c c}
\toprule
  & \textbf{GSM8K} & \textbf{SVAMP} & \textbf{AQuA} & \textbf{MultiArith} & \textbf{AddSub} & \textbf{SingleEq} \\ \midrule
Zero-shot CoT & 75.33 & 77.27 & 58.27 & 96.40 & 87.26 & 92.00 \\ 
Self-correct & 74.13 & 76.40 & 53.67 & 93.33 & 82.62 & 89.13 \\ 
Multi-agent Majority & 81.80 & 79.80 & 62.20 & 97.80 & 89.37 & 93.60 \\ 
Multi-agent Debate & 81.60 & 78.40 & 62.60 & 97.20 & 88.10 & 93.80 \\ \midrule
Ours & \textbf{83.20} & \textbf{83.60} & \textbf{65.35} & \textbf{98.00} & \textbf{89.87} & \textbf{94.20} \\ 
Ours (w/o confidence) & 82.60 & 82.60 & 63.78 & 97.60 & 88.61 & 93.40 \\  
Ours (w/o solution) & 82.80 & 83.30 & 64.96 & 97.40 & \textbf{89.87} & 94.00 \\
\bottomrule
\end{tabular}
\caption{Accuracy of our collaboration strategy compared with previous state-of-the-art methods on six mathematical reasoning datasets. The best results are in \textbf{bold}.}
\label{tab:math_accuracy}
\end{table*}

\begin{table}[t]
\small
\centering
\begin{tabular}{lcc}
\toprule
  & \textbf{ARC-c} & \textbf{StrategyQA} \\ \midrule
Zero-shot CoT & 86.07 & 66.13 \\ 
Self-correct & 57.73 & 52.93 \\ 
Multi-agent Majority & 87.60 & 68.20 \\ 
Multi-agent Debate & 87.40 & 69.40 \\ \midrule
Ours & 88.40 & \textbf{69.80} \\ 
Ours (w/o confidence) & \textbf{88.60} & 69.40 \\ 
Ours (w/o solution) & 88.00 & 67.00 \\
\bottomrule
\end{tabular}
\caption{Accuracy on commonsense reasoning datasets.}
\label{tab:cs_accuracy}
\end{table}

\subsection{Benchmarks}

Our method is evaluated on ten benchmark datasets from three categories of reasoning problems:
mathematical reasoning, commonsense reasoning, and symbolic reasoning.
Table~\ref{tab:statistics} shows the statistics of evaluation datasets.

\noindent \textbf{Mathematical reasoning:}
\begin{itemize}
    \item GSM8K~\cite{cobbe2021gsm}: A high quality linguistically diverse grade-school math word problems created by human problem writers.
    \item SVAMP~\cite{patel-etal-2021-svamp}: A challenging math word problem set with grade level up to 4, by applying simple variations over an existing dataset.
    \item AQuA~\cite{ling-etal-2017-aqua}: A math word problem set with natural language rationales.
    \item MultiArith~\cite{roy-roth-2015-multiarith}: A set of math word problems requiring multiple reasoning steps and operations.
    \item AddSub~\cite{hosseini-etal-2014-addsub}: A set of addition and subtraction math word problems.
    \item SingleEq~\cite{koncel-kedziorski-etal-2015-singleeq}: A single-equation grade-school math word problems with multiple math operations over non-negative rational numbers and one variable.
\end{itemize}

\noindent \textbf{Commonsense reasoning:}
\begin{itemize}
    \item ARC-c~\cite{clark2018arc}: A subset of the AI2 Reasoning Challenge, consisting of challenging science questions that require reasoning and a wide breadth of knowledge to answer the multiple-choice problems correctly.
    \item StrategyQA~\cite{geva-etal-2021-strategyqa}: A commonsense QA task with Yes or No answer format that requires models to perform multi-hop reasoning to answer the questions.
\end{itemize}

\noindent \textbf{Symbolic reasoning:}
\begin{itemize}
    \item Colored Objects~\cite{suzgun-etal-2023-bbh}: Given a context, answer a simple question about the color of an object on a surface.
    \item Penguins~\cite{suzgun-etal-2023-bbh}: Given a unique table of penguins, answer a question about the attributes of the them.
\end{itemize}

\subsection{Baselines}

We compare our peer review collaboration with the following single-agent and multi-agent methods:
\begin{itemize}
    \item Zero-shot CoT~\cite{kojima2022zeroshotcot} utilizes prompts to elicit both reasoning chains and answers from an LLM.
    \item Self-correct~\cite{huang2023noselfcorrect} enables an LLM to iteratively review, find problems, and refine its answer. 
    \item Multi-agent Majority~\cite{wang2023selfconsistency} selects the most consistent answer from several reasoning chains.
    \item Multi-agent Debate~\cite{du2023improving}: Each agent in the group sees solutions of others, updates its own solution, and repeats this process iteratively.
\end{itemize}

\subsection{Implementation Details}
\label{sec:implement}


We use the \texttt{gpt-3.5-turbo-0613} OpenAI API for all main experiments.
In further analysis involving different LLMs for collaboration, we also incorporate the use of \texttt{gpt-3.5-turbo-16k-0613}, \texttt{claude-instant-1.2} and \texttt{claude-2.1}. The details of prompts and hyperparameter settings are in Appendix~\ref{sec:appendix_implement}.

\begin{table}[t]
\small
\centering
\begin{tabular}{lcc}
\toprule
  & \textbf{Colored Objects} & \textbf{Penguins} \\ \midrule
Zero-shot CoT & 66.13 & 70.78 \\ 
Self-correct & 28.27 & 47.95 \\ 
Multi-agent Majority & 70.40 & 73.29 \\ 
Multi-agent Debate & 69.60 & 76.71 \\ \midrule
Ours & 73.20 & \textbf{79.45} \\ 
Ours (w/o confidence) & \textbf{74.40} & \textbf{79.45} \\ 
Ours (w/o solution) & 71.60 & 76.71 \\
\bottomrule
\end{tabular}
\caption{Accuracy on symbolic reasoning datasets.}
\label{tab:sym_accuracy}
\end{table}

\begin{figure*}[ht]
    \centering
    \begin{subfigure}{0.48\textwidth}
        \centering
        \includegraphics[width=\linewidth]{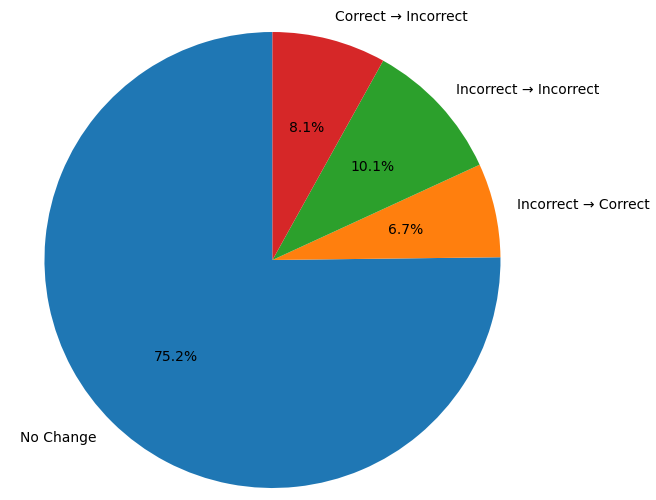} 
        \caption{Self-correct}
    \end{subfigure}
    \begin{subfigure}{0.48\textwidth}
        \centering
        \includegraphics[width=\linewidth]{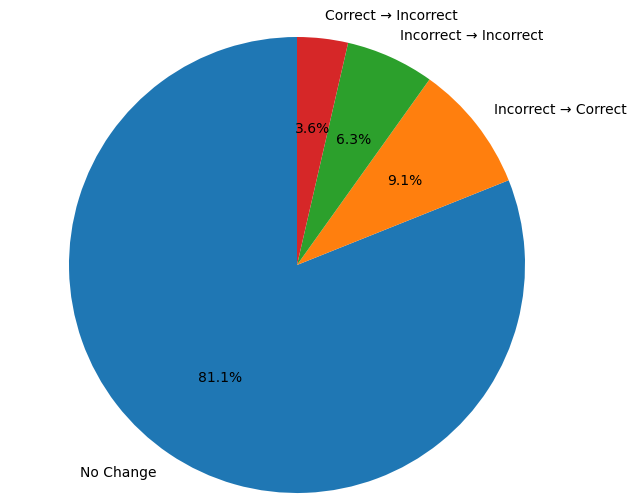} 
        \caption{Our method}
    \end{subfigure}

    \caption{Comparison of the changes in answers on GSM8K after self-correction and our collaboration method.
    \textit{No change}: The answer remains unchanged;
    \textit{Correct$\rightarrow$Incorrect}: A correct answer is changed to an incorrect one;
    \textit{Incorrect$\rightarrow$Correct}: An incorrect answer is revised to a correct one;
    \textit{Incorrect$\rightarrow$Incorrect}: An incorrect answer is altered but remains incorrect.
    }
    \label{fig:gsm8k_changes}
\end{figure*}

\subsection{Results}
\label{sec:results}

\textbf{Main results.}
Table~\ref{tab:math_accuracy} shows the accuracy comparison on the mathematical reasoning datasets.
Our peer review strategy surpasses all single-agent and multi-agent baselines across all datasets,
especially significant in more challenging benchmarks (+1.40\% on GSM8K, +3.80\% on SVAMP, +2.75\% on AQuA over the prior state-of-the-art).

Table~\ref{tab:cs_accuracy} and Table~\ref{tab:sym_accuracy} show the results on commonsense reasoning and symbolic reasoning datasets, respectively.
Our collaboration method consistently outperforms all prior baselines across all datasets.
The effectiveness of self-correction notably decreases in non-math reasoning tasks, 
as seen with a substantial drop from 66.13\% to 28.27\% in accuracy on Colored Objects. 
However, when we introduce feedback from other agents, there is a significant enhancement in performance, rising from 66.13\% to 73.20\%. This demonstrates the advantage of integrating external feedback over relying solely on internal feedback mechanisms.

\noindent \textbf{Ablation study.} 
For the ablation of \textbf{confidence} in reviews,
we alter the prompt of Stage 2 described in Section~\ref{sec:method}.
Specifically, we remove the prompt of eliciting confidence, simplifying it to: 
\textit{``Please examine this agent's reasoning process step by step and offer feedback on its reasoning''}.
Table~\ref{tab:math_accuracy} shows the effectiveness of expressing uncertainty and integrating confidence in feedback on mathematical reasoning datasets.
An example of confidence ablation is shown in Appendix~\ref{sec:confidence_ablation}.
As stated in the prompt of Stage 3, the updated answer uses other agents' solutions and feedback as additional information.
In the absence of confidence scores, the updated answer tends to be misled by ineffective feedback and remains incorrect.
In contrast, the integration of confidence score enables the model to selectively aggregate valuable information.
By focusing more on solutions with higher confidence levels rather than on less reliable feedback, the confidence scores guide the model towards the correct answer.

For commonsense reasoning in Table~\ref{tab:cs_accuracy} and symbolic reasoning in Table~\ref{tab:sym_accuracy},
incorporating confidence can result in unchanged or even reduced performance.
This decrease may stem from the tendency of LLMs to exhibit overconfidence and significant miscalibration when verbalizing their confidence in non-math datasets~\cite{xiong2023uncertain}.
We will further discuss the relationship between the level of confidence and the correctness of feedback in Section~\ref{sec:analysis}.

For the ablation of \textbf{solutions},
we leave out the the dialogue history of Stage 2 during the answer revision process in Stage 3,
so that the peer solutions cannot be seen.
We also change the prompt of Stage 3 into \textit{``Using other agents' feedbacks as additional information, can you provide your answer to the problem? ''}.
From Table~\ref{tab:math_accuracy},~\ref{tab:cs_accuracy} and~\ref{tab:sym_accuracy}, omitting solutions resulted in decreased performance in 9 out of 10 datasets,
indicating that solutions contribute valuable complementary information to the feedback received.
In addition, 
\textit{Ours (w/o solution)}, which involves exchanging only peer feedback,
outperforms \textit{Multi-agent Debate}, which exchanges only peer solutions,
in 8 out of 10 datasets.
This demonstrates that peer feedback is more effective than peer solutions in multi-agent collaboration.

\begin{figure*}[ht]
    \centering
    \begin{subfigure}{0.48\textwidth}
        \centering
        \includegraphics[width=\linewidth]{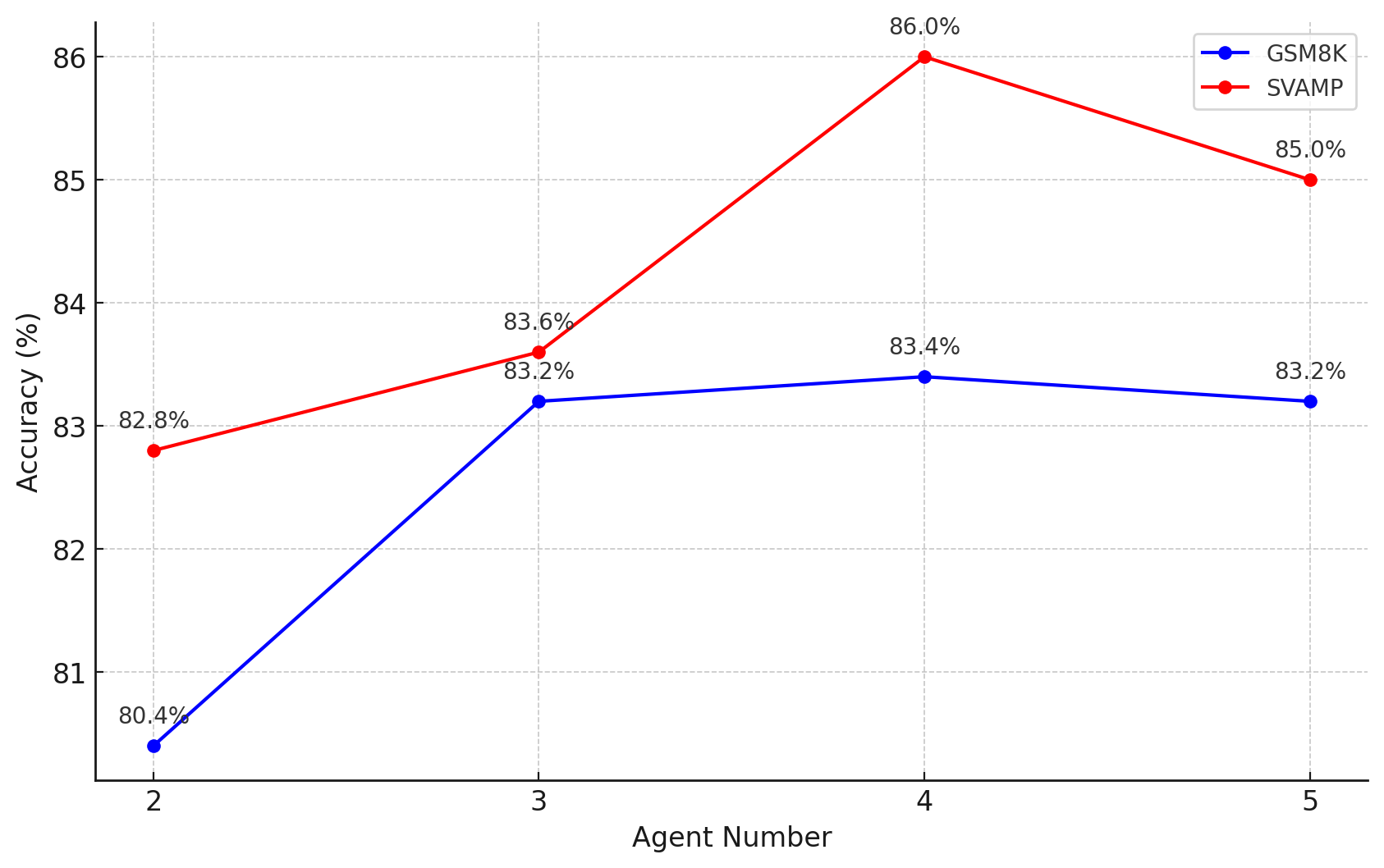} 
        \caption{Accuracy vs. Agent Number}
        \label{fig:agent_number}
    \end{subfigure}
    \hfill 
    \begin{subfigure}{0.48\textwidth}
        \centering
        \includegraphics[width=\linewidth]{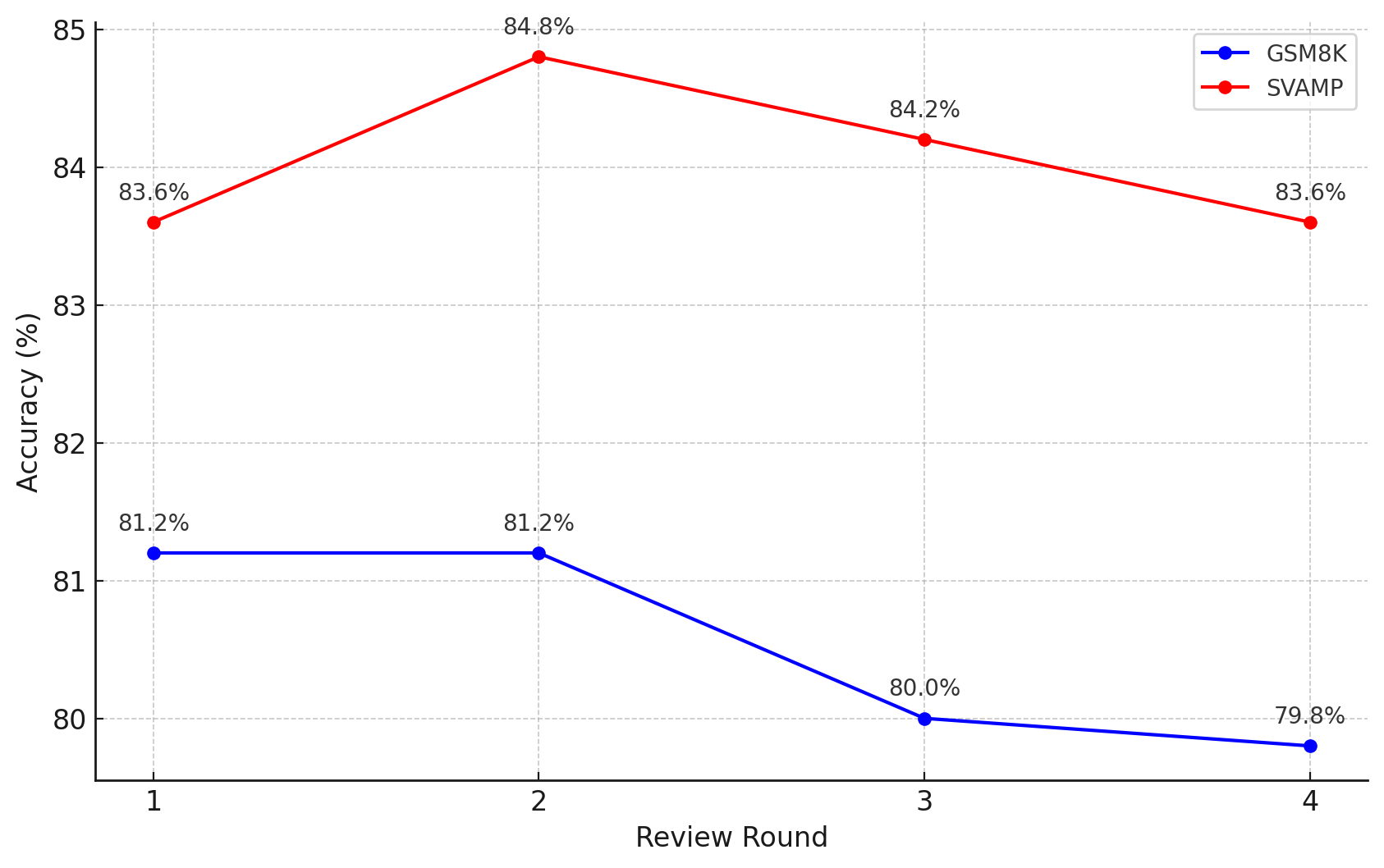} 
        \caption{Accuracy vs. Review Round}
        \label{fig:review_round}
    \end{subfigure}

    \caption{Performance on GSM8K and SVAMP with increased agents and review rounds: Performance shows an upward trend as the number of involved agents increases, while no significant improvement trend is detected with more review rounds.}
    \label{fig:combined_metrics}
\end{figure*}

\begin{figure*}[ht]
    \centering
    \begin{subfigure}[b]{0.45\linewidth}
        \centering
        \includegraphics[width=\linewidth]{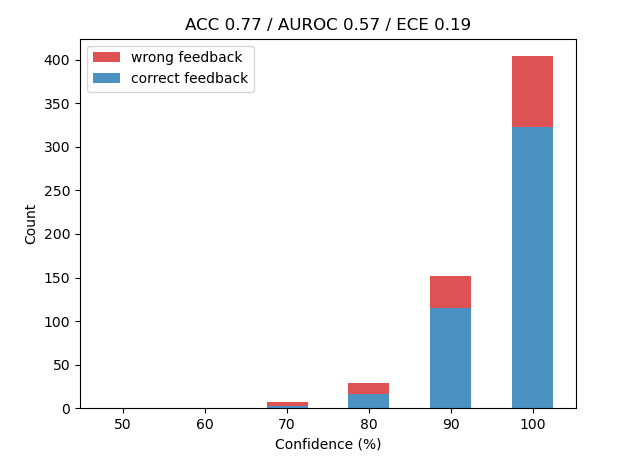}
    \end{subfigure}
    \hfill
    \begin{subfigure}[b]{0.45\linewidth}
        \centering
        \includegraphics[width=\linewidth]{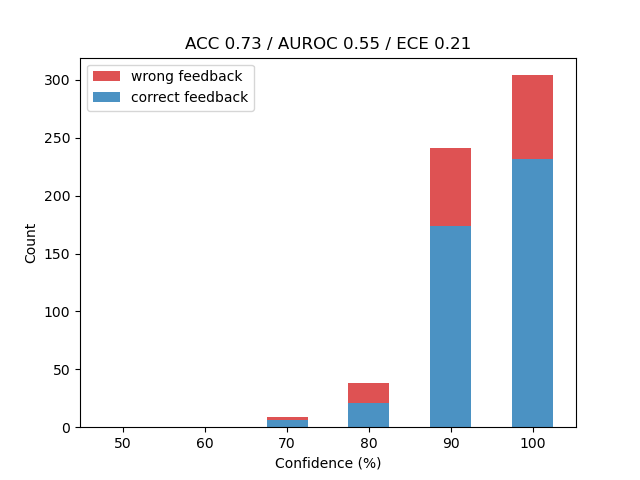}
    \end{subfigure}
    \newline
    \begin{subfigure}[b]{0.45\linewidth}
        \centering  
        \includegraphics[width=\linewidth]{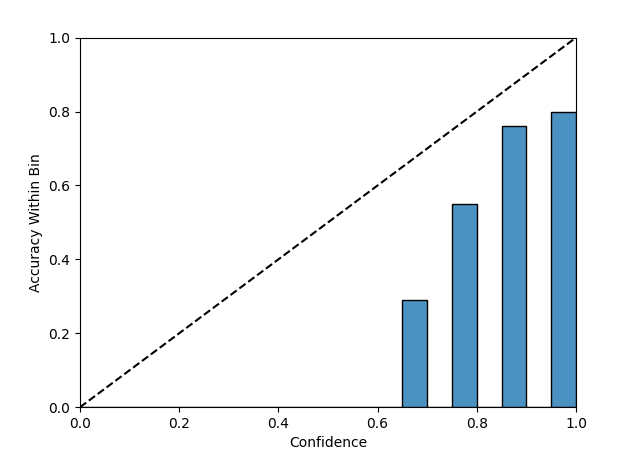}
        \caption{GSM8K}
    \end{subfigure}
    \hfill
    \begin{subfigure}[b]{0.45\linewidth}
        \centering
        \includegraphics[width=\linewidth]{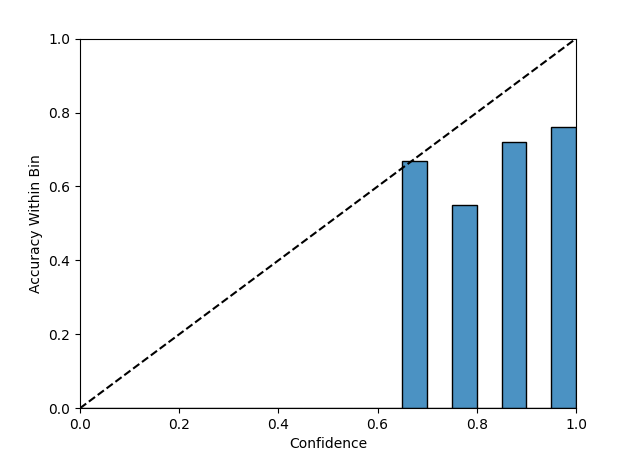}
        \caption{Penguins}
    \end{subfigure}
    \caption{\textbf{First row}: Distribution of verbalized confidence on GSM8K and Penguins. Most samples fall into the 80\% to 100\% range.
    To better illustrate the major confidence distribution, we set the minimal confidence threshold at 50\%, as very few confidences fall below.
    \textbf{Second row}: Reliability diagram on GSM8K and Penguins. The accuracy within each bin is much lower than its corresponding confidence, showing significant overconfidence.
    }
    \label{fig:confidence_acc}
\end{figure*}

\begin{table*}[ht]
\small
\centering
\begin{tabular}{lcccccc}
\toprule
\textbf{Peer Review} & \textbf{Capability Gap} & \textbf{Diversity} & \textbf{LLM} & \textbf{Initial Acc.} & \textbf{Updated Acc.} & \textbf{$\Delta$} \\
\midrule
\multirow{2}{*}{\makecell[l]{\texttt{gpt-3.5-turbo-0301}\\ \& \texttt{gpt-3.5-turbo-0613}}} & \multirow{2}{*}{2.8} & \multirow{2}{*}{35.4} & \texttt{gpt-3.5-turbo-0301} & 72.6 & 80.0 & \green{+7.4} \\
 & & & \texttt{gpt-3.5-turbo-0613} & 75.4 & 77.8 & \green{+2.4} \\ \midrule
\multirow{2}{*}{\makecell[l]{\texttt{gpt-3.5-turbo-0613}\\ \& \texttt{claude-instant-1.2}}} & \multirow{2}{*}{10.0} & \multirow{2}{*}{31.4} & \texttt{gpt-3.5-turbo-0613} & 75.4 & 83.0 & \green{+7.6} \\
 & & & \texttt{claude-instant-1.2} & 85.4 & 86.0 & \green{+0.6} \\ \midrule
\multirow{2}{*}{\makecell[l]{\texttt{claude-instant-1.2}\\ \& \texttt{claude-2.1}}} & \multirow{2}{*}{2.2} & \multirow{2}{*}{15.8} & \texttt{claude-instant-1.2} & 85.4 & 86.6 & \green{+1.2} \\
 & & & \texttt{claude-2.1} & 87.6 & 87.4 & \textcolor{red}{-0.2} \\
\bottomrule
\end{tabular}
\caption{The peer review collaboration results with different LLMs. LLMs with smaller capability gap but higher diversity can improve and collaborate with each other better.}
\label{tab:cap_vs_diversity}
\end{table*}

\subsection{Analysis} 
\label{sec:analysis}

\textbf{Comparison of answer changes.} 
We compare the revised solutions with the initial solutions, and summarize the results of changes in answers after self-correction and our method.
Figure~\ref{fig:gsm8k_changes} and Appendix~\ref{sec:appendix_ans_change} show the answer changes on GSM8K and Penguins respectively. 
When applying self-correction on GSM8K, the model is more likely to modify a correct answer to an incorrect one than to revise an incorrect answer to a correct one.
In the non-math dataset Penguins,
there is a higher chance that the model alters its answer after self-correction. 
In 37.0\% of the time, the model changes a correct answer into an incorrect one.
However, our method exhibits a different pattern for both the GSM8K and Penguins datasets, with the portion of \textit{Incorrect$\rightarrow$Correct} significantly larger than \textit{Correct$\rightarrow$Incorrect}.
Both figures illustrate the benefits of incorporating external feedback over relying solely on internal feedback mechanisms.

\noindent \textbf{Impact of agent numbers and review rounds.} 
We study how the number of agents and review rounds affects performance.
From Figure~\ref{fig:combined_metrics}a, 
an upward trend is observed in the relationship between the accuracy and the agent number.
The growth pattern reaches an apex at 4 agents before declining at agent number 5.
Regarding the relationship between the accuracy and the review rounds,
as shown in Figure~\ref{fig:combined_metrics}b,
no significant improvement trend is detected with more review rounds.
This observation is consistent with the findings of~\citet{chan2023chateval},
which suggest that continual discussions often lead to reduced performance.
The decrease may be due to issues associated with the ever-increasing long-context modeling, which negatively affects the performance.

\noindent \textbf{Interplay between confidence scores and feedback correctness.}
We manually annotate the correctness of 600 feedback each from the GSM8K and Penguins datasets.
We present the relationship between confidence scores and feedback correctness in Figure~\ref{fig:confidence_acc}, along with the evaluation metrics such as feedback accuracy, Area Under the Receiver Operating Characteristic curve (AUROC) and Expected Calibration Error (ECE).
We have the following observations:
(1) From the confidence distribution in the first row of Figure~\ref{fig:confidence_acc}, most verbalized confidence scores fall within the 80\% to 100\% range.
(2) The second row of Figure~\ref{fig:confidence_acc} shows that the accuracy of each bar ranging from 80\% to 100\% is much less than 80\%,
suggesting significant overconfidence and miscalibration in LLMs.
(3) When comparing the math dataset GSM8K with the non-math dataset Penguins, the feedback in the non-math dataset tends to be more poorly calibrated and exhibit greater overconfidence. 
This observation might explain the finding in ablation study (Section~\ref{sec:results}), which suggests that incorporating confidence on non-math datasets may not improve performance.

\noindent \textbf{Performance variability with diverse LLMs.}
We explore what kind of LLMs can collaborate well and improve each other through the peer review process.
We set the agent number at 2, and use GSM8K for the analysis.
We focus on two key factors: the capability gap and diversity between LLMs.
The capability gap refers to the difference in the accuracy of initial responses provided by two LLMs.
The diversity is measured by \texttt{INCON}~\cite{xiong-etal-2023-examining}. Specifically, suppose there are $n$ LLMs $L=\{l_1,\cdots,l_n\}$, and a dataset with $m$ samples $X=\{x_1,\cdots,x_m\}$. We define $p^i_j$ as the prediction of $l_i$ on $x_j$. Then the \texttt{INCON} of $L$ on $X$ can be defined as:
\begin{equation}
\small
    \texttt{INCON}=\sum_{k=1}^m\frac{\Phi(p^1_k,\cdots,p^n_k)}{m},
\end{equation}
$\Phi$ is a sign function, it will be assigned a value of 1 if there are any two variables in $\Phi$ that are not equal, otherwise, $\Phi$ takes a value of 0.

The collaboration results are presented in Table~\ref{tab:cap_vs_diversity}.
We have the following conclusions:
(1) More powerful models provide more effective feedback. 
The \texttt{0301} version of \texttt{gpt-3.5-turbo} contributes to a 2.4\% accuracy improvement in the \texttt{0613} version, while the more advanced \texttt{claude-instant-1.2} leads to a significant 7.6\% increase.
(2) LLMs with smaller capability gap can collaborate better. 
By comparing the first two rows, 
despite similar levels of diversity, a capability gap increase from 2.8 to 10.0 corresponds to a decrease in performance enhancement for the stronger model, from a 2.4\% to just a 0.6\% improvement.
(3) LLMs with higher diversity can collaborate better.
With a similar capability gap, the collaboration between \texttt{gpt-3.5-turbo-0301} and its \texttt{0613} version, which have considerable diversity, results in mutual improvement. In contrast, two \texttt{claude} models, with substantially lower diversity, show that the stronger model does not benefit from the collaboration.

\noindent \textbf{Impact of agent roles.}
Our default experiments do not include any system messages (i.e. role specification).
Here we explore diverse role prompts for agent initialization.
The roles are automatically generated by SPP~\cite{wang2023unleashing} and detailed in Appendix~\ref{sec:appendix_roles}.
Experiment results in Appendix~\ref{sec:appendix_roles} highlight the cruciality of diverse role prompt design in multi-agent collaboration.

\subsection{Case study}
\label{sec:cases}
In this section, we present the complete peer review collaboration process to a math question in GSM8K~\cite{cobbe2021gsm}.
As shown in Appendix~\ref{sec:appendix_cases}, the procedure begins with two incorrect initial answers. 
Through a series of dynamic, human-like discussions and critiques among the agents
(for example, Agent B explicitly addresses Agent A's misunderstanding of the question and the mistake in the ratio step),
all three agents eventually converge on the same correct solution.

\section{Conclusion}

In this work, we let a single model think outside the box by engaging multiple models to collaborate on complex reasoning.
We introduce a multi-agent collaboration strategy that emulates the academic peer review process,
following an iterative cycle of creation, review, and revision.
Extensive experiments show that our approach delivers superior accuracy across all ten datasets and three tasks compared to existing methods.
Further analysis shows the effectiveness of confidence integration and
demonstrates the superiority of feedback exchange over mere solution sharing.


\section*{Limitations}

Despite the effectiveness of multi-agent methods,
it is essential to consider the associated costs, particularly in terms of token consumption. 
The balance between effectiveness and efficiency emerges as a critical area for exploration.
In addition, the underlying factors for LLM collaboration are more than just capability and answer diversity.
The factors also include role and tool specification of agents.
Further research may focus on identifying which types of agents can collaborate more effectively together, thereby providing deeper insights into the optimization of multi-agent systems in practical applications.

\bibliography{custom}


\appendix

\section{Implementation Details}
\label{sec:appendix_implement}

We use the \texttt{gpt-3.5-turbo-0613} OpenAI API for all experiments, with all parameters in default setting.
We fix the agent number as 3 and the round number of review cycle as 1.
We also apply \texttt{gpt-3.5-turbo-16k-0613} to explore scenarios with a greater number of agents and additional review rounds in Section~\ref{sec:analysis}.
Specifically, we utilize the \texttt{16k} version for experiments when the total count of agents is set to 5.
We also use the \texttt{16k} version for an increased number of review cycles. 
The review is conducted over four rounds, with accuracy being calculated after each round, 
so the result at the first round in Figure~\ref{fig:combined_metrics} may differ from the reported result in Table~\ref{tab:math_accuracy}.

Our implementation of the \textit{self-correct} prompt draws from \citet{huang2023noselfcorrect}.
For the implementation of \textit{multi-agent debate}~\cite{du2023improving},
we utilize the official Github repository\footnote{The code for \citet{du2023improving} is at  \url{https://github.com/composable-models/llm_multiagent_debate}}. 
This implementation adheres to the default configuration of three agents and two debate rounds.

For evaluation metric,
we follow~\citet{kojima2022zeroshotcot} and report the accuracy of all methods across datasets.
Due to the rate limits and a restricted budget,
our experiments use a maximum of 500 samples, randomly selected from each dataset.

\section{Example of Confidence Ablation}
\label{sec:confidence_ablation}

Here we present an example of confidence ablation in Table~\ref{tab:example_conf_ablation}.
The initial solution of Agent A is wrong.
In the absence of confidence scores, the final answer is misled by ineffective feedback and remains unchanged.
In contrast, the integration of confidence score enables the model to focus more on solutions with higher confidence levels rather than on less reliable feedback, eventually guiding the model towards the correct answer.

\begin{table*}[ht]
\small
\centering
\begin{tabular}{p{0.07\textwidth} p{0.35\textwidth} p{0.49\textwidth}}
\toprule
\multirow{2}{0.07\textwidth}{\textbf{Example}} & \multicolumn{2}{p{0.86\textwidth}}{\textbf{Q}: If there are 10 eggs in a basket, and there are twice as many eggs in a second basket, how many eggs are in both baskets put together?} \\
\cmidrule{2-3}
                         & \multicolumn{2}{p{0.84\textwidth}}{\textbf{A}: 10 + 10 * 2 = \green{30}} \\
\cmidrule{1-3}
\textbf{Stage 1: Create}                  & \multicolumn{2}{p{0.86\textwidth}}{Let's call the number of eggs in the first basket x. We know that \red{x + 2x = 10}, since the total number of eggs in both baskets is given as 10. x = 10/3. Since we need a whole number of eggs, we can round this value down to the nearest whole number, which is 3. Therefore, there are x = 3 eggs in the first basket and 2x = 2(3) = 6 eggs in the second basket. Hence, there are \red{9} eggs in both baskets put together. } \\
\cmidrule{1-3}
\multirow{4}{0.07\textwidth}{\textbf{Stage 2: Review}} & \multicolumn{2}{p{0.84\textwidth}}{\textit{Receives a \green{correct} solution from Agent B.}} \\
\cmidrule{2-3}
                         & Overall, the agent's reasoning and steps are \green{accurate and correct}.                     & I would rate my confidence in the feedback given a \textbf{10}. Overall, the agent's reasoning and steps are \green{accurate and correct}.                    \\
\cmidrule{2-3}
                         & \multicolumn{2}{p{0.8\textwidth}}{\textit{Receives another \green{correct} solution from Agent C.}} \\
\cmidrule{2-3}
                         & The agent's reasoning and steps are \green{accurate and correct}.                     & I would rate my confidence in the feedback given a \textbf{10}. The agent's reasoning and steps are \green{accurate and correct}.                    \\
\cmidrule{1-3}
\multirow{2}{0.07\textwidth}{\textbf{Stage 3: Revise}}  & \textit{Receives two \red{wrong} feedback from Agent B and Agent C: The agent's reasoning process is \red{correct}, and each step is logical and accurate; The agent's reasoning was mostly \red{correct except for the rounding of the value of x.}} & \textit{Receives two \red{wrong} feedback from Agent B and Agent C: The agent's reasoning process is \red{correct}, and each step is logical and accurate. My confidence in this feedback is \textbf{9}; The agent's reasoning was mostly \red{correct except for the rounding of the value of x.} I would rate my confidence in this feedback as \textbf{7} out of 10.} \\
\cmidrule{2-3}
                         & x + 2x = 10. x = 3. The answer is 3 + 3 * 2 = \red{9} eggs in both baskets put together.                     & 10 + 10 * 2 = 30. The correct answer is \green{30} eggs in both baskets put together.                    \\
\bottomrule
\end{tabular}
\caption{The responses of Agent A without confidence (left) and with confidence (right). The correct answer and feedback are in \green{green}, and the wrong ones are in \red{red}. The confidence score assists Agent A towards the correct answer. For ease of read, only the key points of the responses are displayed.}
\label{tab:example_conf_ablation}
\end{table*}

\section{Answer Change Comparison}
\label{sec:appendix_ans_change}

\begin{figure*}[ht]
    \centering
    \begin{subfigure}{0.48\textwidth}
        \centering
        \includegraphics[width=\linewidth]{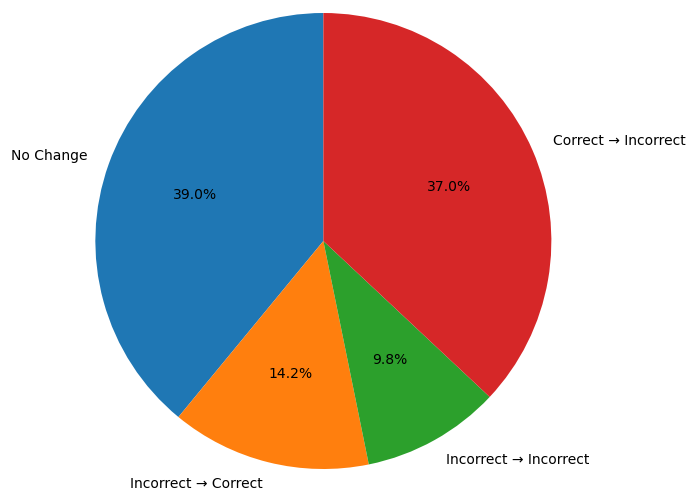} 
        \caption{Self-correct}
    \end{subfigure}
    \begin{subfigure}{0.48\textwidth}
        \centering
        \includegraphics[width=\linewidth]{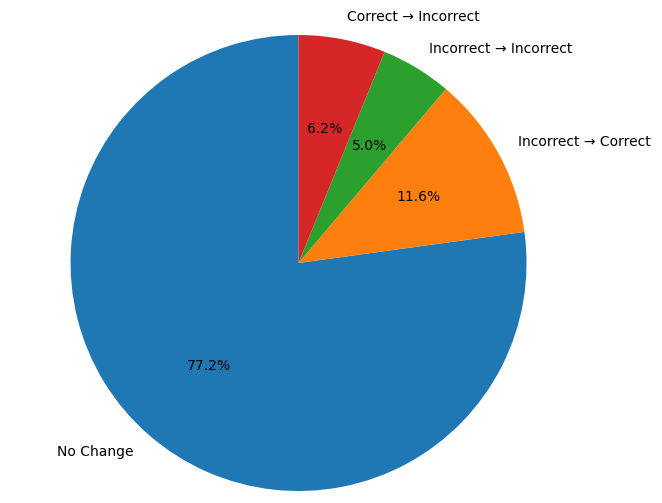} 
        \caption{Our method}
    \end{subfigure}

    \caption{Comparison of the changes in answers on Penguins after self-correction and our collaboration method.
    \textit{No change}: The answer remains unchanged;
    \textit{Correct$\rightarrow$Incorrect}: A correct answer is changed to an incorrect one;
    \textit{Incorrect$\rightarrow$Correct}: An incorrect answer is revised to a correct one;
    \textit{Incorrect$\rightarrow$Incorrect}: An incorrect answer is altered but remains incorrect.}
    \label{fig:penguins_changes}
\end{figure*}

We compare the revised answers with the initial answers, and summarize the results of changes in answers after self-correction and our method.
Figure~\ref{fig:penguins_changes} show the answer changes on the Penguins dataset.

\section{Impact of Diverse Role Prompts}
\label{sec:appendix_roles}

We conduct experiments on GSM8K and SVAMP.
The following five roles are automatically generated through SPP~\cite{wang2023unleashing}:

\noindent \textbf{AI Assistant.} \textit{You are an advanced AI capable of performing complex calculations, logical reasoning, and providing efficient solutions to mathematical problems.}

\noindent \textbf{Math Teacher.} \textit{You are an experienced math teacher with a deep understanding of mathematical concepts, methods, and pedagogy. You can offer insights into problem-solving strategies and educational approaches to mathematics.}

\noindent \textbf{Mathematical Scientist.} \textit{You are a scientist who specializes in advanced mathematical theories and research. You can contribute deep theoretical knowledge and expertise in cutting-edge mathematical areas, such as topology, abstract algebra, or number theory.}

\noindent \textbf{Engineer.} \textit{You are a professional engineer with practical experience in applying mathematical concepts to solve real-world problems. You can offer a practical viewpoint on how mathematics is used in designing, analyzing, and improving systems and structures.}

\noindent \textbf{Computer Scientist.} \textit{You are an expert in computer science, well-versed in algorithms, computational theory, and programming. You can assist in solving problems that involve algorithmic thinking, computational complexity, or numerical methods.}

We study whether it is necessary to design diverse role prompts for multi-agent collaboration.
Table~\ref{tab:role} shows the experiment results.
In the \textit{single role} setting, all agents are AI assistants.
In the \textit{diverse role} setting, for every question, agent roles are randomly selected from the above 5 roles.
The results illustrate that our method with the same role prompt underperforms that with diverse role prompt,
highlighting the cruciality of diverse role specification in multi-agent collaboration.

\begin{table}[ht]
\centering
\begin{tabular}{lcc}
\toprule
 & \textbf{GSM8K} & \textbf{SVAMP} \\
\midrule
Single Agent & 75.33 & 77.27 \\
Multi Agent (single role) & 81.80 & 84.80 \\
Multi Agent (diverse role) & \textbf{82.80} & \textbf{85.40} \\
\bottomrule
\end{tabular}
\caption{Effects of diverse role specification on GSM8K and SVAMP.}
\label{tab:role}
\end{table}


\section{Case Study}
\label{sec:appendix_cases}


The process unfolds across three stages, each illustrated in separate tables:
Table~\ref{tab:stage1} illustrates the first ``creation'' stage.
Table~\ref{tab:stage2AB} and Table~\ref{tab:stage2C} both show the second ``review'' stage.
Finally, the ``revision'' stage is detailed in Table~\ref{tab:stage3}.

\begin{table*}[p]
\centering
\small
\begin{tabular}{rp{0.85\textwidth}}
\toprule

\multirow{9.5}{*}{\textbf{Example}}

    & \textbf{Q}: Elise has been selling her Dad's collection of 250 books for three years. Each book sells at 20\$, and she sold twice as many books in the first year as she has sold in the current year. There are currently 50 unsold books, and her sales number this year is 45. What's the total amount of money she earned in the second year?  (later denoted as \texttt{Question})    \\

\cmidrule{2-2}

    & \textbf{A}: If her sales in the current year are 45, she sold 45*2 = <<45*2=90>>90 books in the first year. The total number of books sold in the first and current year is 90+45 = <<90+45=135>>135. The total number of books sold in the second year is 250 of the original number of books - 135 sold books - 50 unsold books = <<250-135-50=65>>65 books. If each book sells at 20\$, then she earned 20*65 = \$<<20*65=1300>>1300 in the second year \#\#\#\# \textbf{1300}  \\

\cmidrule{1-2}

\multirow{12.5}{*}{\textbf{Agent A}}

    & \textbf{Q}: Can you solve the following math problem? \texttt{\{Question\}} Explain your reasoning. Your final answer should be a single numerical number, in the form \textbackslash boxed\{answer\}, at the end of your response.  \\ 

\cmidrule{2-2}

    & \textbf{A}: Let's break down the information given: Elise sold twice as many books in the first year as she has sold in the current year. \red{If she sold 45 books this year, then she sold 45/2 = 22.5 books in the first year.} However, since we cannot have half a book, we'll assume that she actually sold 23 books in the first year. There are currently 50 unsold books, which means that she has sold a total of 250 - 50 = 200 books in the past three years. If she sold 23 books in the first year and 45 books this year, then she sold a total of 23 + 45 = 68 books in the past two years. To calculate the number of books she sold in the second year, we subtract the number of books sold in the first year and this year from the total number of books sold in the past two years: 68 - 23 - 45 = 68 - 68 = 0. Therefore, Elise did not sell any books in the second year. Since each book sells for \$20, she did not earn any money in the second year. Thus, the total amount of money she earned in the second year is \textbackslash boxed\{\red{0}\}. (later denoted as \texttt{Answer A}) \\

\cmidrule{1-2}

\multirow{13.5}{*}{\textbf{Agent B}}

    & \textbf{Q}: Can you solve the following math problem? \texttt{\{Question\}} Explain your reasoning. Your final answer should be a single numerical number, in the form \textbackslash boxed\{answer\}, at the end of your response.  \\ 

\cmidrule{2-2}

    & \textbf{A}: Let's start by finding out how many books Elise sold in the first year. We are given that she sold twice as many books in the first year as she has sold this year. Since she has sold 45 books this year, she sold 2 * 45 = 90 books in the first year. In the first year, she sold a total of 90 books * \$20 = \$<<90*20=1800>>1800. Now let's find out how many books Elise sold in the current year. We are given that she sold 45 books this year and there are currently 50 unsold books. Therefore, in the current year, she had sold a total of 45 books + 50 unsold books = 95 books. Since we know Elise sold a total of 250 books over three years, and we already calculated the number of books sold in the first and current years, we can find out the number of books she sold in the second year. The total number of books sold in the second year is 250 books - 90 books - 95 books = 65 books. We know that each book sells for \$20, so in the second year, Elise earned a total of 65 books * \$20 = \$<<65*20=1300>>1300. Therefore, Elise earned a total of \$1300 in the second year. Our final answer is \textbackslash boxed\{\green{1300}\}. (later denoted as \texttt{Answer B})    \\

\cmidrule{1-2}

\multirow{15.5}{*}{\textbf{Agent C}}

    & \textbf{Q}: Can you solve the following math problem? \texttt{\{Question\}} Explain your reasoning. Your final answer should be a single numerical number, in the form \textbackslash boxed\{answer\}, at the end of your response.  \\ 

\cmidrule{2-2}

    & \textbf{A}: Let's set up an equation to represent the information given. Let's call the number of books Elise sold in the current year "x". According to the problem, she sold twice as many books in the first year as she has sold in the current year, so she sold 2x books in the first year. In the three years, Elise sold a total of 250 books, and since she sold x books in the current year and 2x books in the first year, she sold 250 - (x + 2x) = 250 - 3x books in the second year. We are given that there are currently 50 unsold books, which means that the total number of books sold is 250 - 50 = 200. \red{Since the total number of books sold in the second year is equal to 200 - (x + 2x), we can set up the equation: 200 - 3x = 45.} Simplifying this equation, we get: -3x = -155. Dividing by -3 on both sides, we find: x = 155/3. Therefore, Elise sold 155/3 books in the current year. Now, to find the total amount of money she earned in the second year, we need to multiply the number of books sold in the second year by the price per book. The price per book is \$20, and the number of books sold in the second year is 200 - (x + 2x) = 200 - 3x = 200 - 3(155/3) = 200 - 155 = 45 books. So, Elise earned a total of 45 * \$20 = \$<<45*20=900>>900 in the second year. Therefore, Elise earned a total of \$900 in the second year. Answer: \textbackslash boxed\{\red{900}\}. (later denoted as \texttt{Answer C})    \\

\bottomrule
\end{tabular}
\caption{\textbf{Stage 1: Create.} Each agent independently constructs its own solution. The first row ``Example'' shows a question and its gold answer in GSM8K~\cite{cobbe2021gsm}. The initial responses of Agent A and Agent C are wrong, and only Agent B is correct. The wrong answers and the critical wrong steps are in \red{red}, and the correct answer is in \green{green}.} 
\label{tab:stage1}
\end{table*}

\begin{table*}[p]
\centering
\small
\begin{tabular}{lp{0.85\textwidth}}
\toprule

\multirow{24.5}{*}{\textbf{Agent A}}

    & \textbf{Q}: Here is a solution from another agent: \texttt{\{Answer B\}} Please examine this agent's reasoning process step by step and offer feedback on its reasoning. You can rate your confidence in your feedback on a scale from 1-10, where 10 indicates the highest level of confidence.  \\ 

\cmidrule{2-2}

    & \textbf{A}: I would rate my confidence in this agent's reasoning process as a 9. \green{The agent's solution is correct}, and they have provided a clear and logical breakdown of the problem. They correctly calculated that Elise sold 90 books in the first year, based on the information given, and found that she sold a total of 95 books in the current year. Subtracting this from the total number of books sold over the three years, they correctly found that Elise sold 65 books in the second year. Finally, they correctly multiplied this number by the price per book to find that Elise earned \$1300 in the second year. Overall, the agent's reasoning process is well-explained and logical. (later denoted as \texttt{Feedback A $\rightarrow$ B}) \\

\cmidrule{2-2}

    & \textbf{Q}: Here is a solution from another agent: \texttt{\{Answer C\}} Please examine this agent's reasoning process step by step and offer feedback on its reasoning. You can rate your confidence in your feedback on a scale from 1-10, where 10 indicates the highest level of confidence.  \\ 

\cmidrule{2-2}

    & \textbf{A}: I would rate my confidence in this agent's reasoning process as a 7. The agent correctly set up the equation to represent the information given, relating the number of books sold in the current year to the number of books sold in the first and second years. They correctly determined that the total number of books sold is 200, based on the information given about unsold books. However, the agent made a mistake in their calculation of the number of books sold in the second year. They incorrectly substituted the value of x into the equation and found that Elise sold 45 books in the second year. \green{This is inconsistent with the information provided, which states that Elise sold 45 books in the current year, not the second year.} Therefore, the calculation of the total amount of money earned in the second year, \$900, is \green{incorrect}. Overall, the reasoning process is mostly correct, but there is a mistake in the calculation of the number of books sold in the second year. (later denoted as \texttt{Feedback A $\rightarrow$ C}) \\

\cmidrule{1-2}

\multirow{31.5}{*}{\textbf{Agent B}}

    & \textbf{Q}: Here is a solution from another agent: \texttt{\{Answer A\}} Please examine this agent's reasoning process step by step and offer feedback on its reasoning. You can rate your confidence in your feedback on a scale from 1-10, where 10 indicates the highest level of confidence.  \\ 
    
\cmidrule{2-2}

    & \textbf{A}: I rate my confidence in my feedback as a 10. Feedback: The reasoning process provided by the other agent is \green{incorrect}. Let's go through the errors in their solution: The other agent assumes that Elise actually sold 23 books in the first year, even though it is specified that she sold twice as many books in the first year as she has sold in the current year. Since she sold 45 books this year, \green{the number of books sold in the first year should be 2 * 45 = 90, not 23.} Similarly, the calculation of the total number of books sold in the past two years is incorrect. It should be 90 + 45 = 135, not 68. Based on these incorrect calculations, the solution concludes that Elise did not sell any books in the second year. However, this is not accurate because we have not determined the number of books sold in the second year correctly. Therefore, the other agent's reasoning process contains errors and does not provide a correct solution to the math problem. (later denoted as \texttt{Feedback B $\rightarrow$ A})    \\

\cmidrule{2-2}

    & \textbf{Q}: Here is a solution from another agent: \texttt{\{Answer C\}} Please examine this agent's reasoning process step by step and offer feedback on its reasoning. You can rate your confidence in your feedback on a scale from 1-10, where 10 indicates the highest level of confidence.  \\ 

\cmidrule{2-2}

    & \textbf{A}: I rate my confidence in my feedback as a 10. Feedback: The reasoning process provided by the other agent is \red{correct}. Let's go through the steps of their solution: The agent sets up an equation to represent the information given, with "x" representing the number of books Elise sold in the current year. They correctly determine that Elise sold twice as many books in the first year as she has sold in the current year, so the number of books sold in the first year is 2x. The agent then uses the total number of books sold in the three years (250) and subtracts the number of books sold in the current year and the first year to find the number of books sold in the second year, which is correctly represented as 250 - (x + 2x). The total number of books sold is given as 200 (250 - 50), and the agent correctly sets up the equation 200 - 3x = 45 to represent the number of books sold in the second year, with 45 being the number of books sold in the current year. The agent solves the equation correctly, finding x = 155/3 as the number of books sold in the current year. Finally, the agent multiplies the number of books sold in the second year by the price per book (\$20) to find the total amount of money earned in the second year, which is correctly calculated as 45 * \$20 = \$900. Therefore, the other agent's reasoning process is correct and provides a valid solution to the math problem. (later denoted as \texttt{Feedback B $\rightarrow$ C})    \\

\bottomrule
\end{tabular}
\caption{\textbf{Stage 2 of Agent A and B: Review.} Each agent provides reviews on the solutions of others. Agent A accurately assesses the answers of others and correctly identifies the primary error in Agent C's solution. Conversely, while Agent B successfully points out the error of Agent A, it fails to pinpoint the error of Agent C. Incorrect feedback is in \red{red}, and the correct feedback and the accurately identified errors are in \green{green}.} 
\label{tab:stage2AB}
\end{table*}

\begin{table*}[ht!]
\centering
\small
\begin{tabular}{lp{0.85\textwidth}}
\toprule

\multirow{26.5}{*}{\textbf{Agent C}}

    & \textbf{Q}: Here is a solution from another agent: \texttt{\{Answer A\}} Please examine this agent's reasoning process step by step and offer feedback on its reasoning. You can rate your confidence in your feedback on a scale from 1-10, where 10 indicates the highest level of confidence.  \\ 

\cmidrule{2-2}

    & \textbf{A}: The reasoning of the second agent is \green{incorrect}. Here's a breakdown of the errors: The assumption that Elise sold 23 books in the first year is incorrect. The problem states that she sold twice as many books in the first year as she has sold in the current year. Since she sold 45 books this year, \green{it means she sold 45 * 2 = 90 books in the first year}. The calculation of the total number of books sold in the past two years is incorrect. It should be 90 (number of books sold in the first year) + 45 (number of books sold this year) = 135, not 68. The conclusion that Elise did not sell any books in the second year is incorrect. The problem does not provide explicit information about the number of books sold in the second year, so we cannot assume it to be zero. As a result, the calculation of the money earned in the second year as \$0 is also incorrect. Confidence in feedback: 10/10. (later denoted as \texttt{Feedback C $\rightarrow$ A})    \\ 

\cmidrule{2-2}

    & \textbf{Q}: Here is a solution from another agent: \texttt{\{Answer B\}} Please examine this agent's reasoning process step by step and offer feedback on its reasoning. You can rate your confidence in your feedback on a scale from 1-10, where 10 indicates the highest level of confidence.  \\ 

\cmidrule{2-2}

    & \textbf{A}: The reasoning of the third agent is \green{correct}. Here's a breakdown of their solution: They correctly calculate that Elise sold 90 books in the first year, as she sold twice as many books as she did in the current year (45). They then correctly calculate the earnings from the first year by multiplying the number of books sold (90) by the price per book (\$20), giving a total of \$1800. They determine that Elise sold a total of 95 books in the current year (45 books sold + 50 unsold books). They correctly calculate the number of books sold in the second year by subtracting the number of books sold in the first and current years (250 books - 90 books - 95 books = 65 books). They calculate the earnings from the second year by multiplying the number of books sold in the second year (65) by the price per book (\$20), giving a total of \$1300. The agent's reasoning process is sound, and they correctly follow through the steps to arrive at the correct answer. Confidence in feedback: 10/10. 
    (later denoted as \texttt{Feedback C $\rightarrow$ B})    \\

\bottomrule
\end{tabular}
\caption{\textbf{Stage 2 of Agent C: Review.} Agent C accurately assesses the answers of others and correctly identifies the primary error in Agent A's solution. The correct feedback and the accurately identified errors are in \green{green}.} 
\label{tab:stage2C}
\end{table*}

\begin{table*}[ht!]
\centering
\small
\begin{tabular}{rp{0.85\textwidth}}
\toprule

\multirow{14.5}{*}{\textbf{Agent A}}

    & \textbf{Q}: Here are the feedbacks for your solution from the above two agents: One agent feedback: \texttt{\{Feedback B $\rightarrow$ A\}} One agent feedback: \texttt{\{Feedback C $\rightarrow$ A\}} Using other agents' solutions and feedbacks as additional information, can you provide your answer to the math problem? The original math problem is \{Question\} Your final answer should be a single numerical number, in the form \textbackslash boxed\{answer\}, at the end of your response.  \\ 

\cmidrule{2-2}

    & \textbf{A}: Based on the corrected information from the feedback received, let's analyze the problem again: Elise sold twice as many books in the first year as she has sold in the current year. Since she sold 45 books this year, the number of books sold in the first year is 2 * 45 = 90 books. The total number of books sold in the past two years is 90 (number of books sold in the first year) + 45 (number of books sold this year) = 135 books. The total number of books sold in the three years is 250 (total number of books) - 50 (unsold books) = 200 books. Therefore, the number of books sold in the second year is 200 (total number of books sold in two years) - 135 (number of books sold in the past two years) = 65 books. Since each book sells for \$20, the total amount of money Elise earned in the second year is 65 books * \$20 = \$1300. Therefore, the total amount of money Elise earned in the second year is \textbackslash boxed\{\green{1300}\}.  \\

\cmidrule{1-2}

\multirow{7.5}{*}{\textbf{Agent B}}

    & \textbf{Q}: Here are the feedbacks for your solution from the above two agents: One agent feedback: \texttt{\{Feedback A $\rightarrow$ B\}} One agent feedback: \texttt{\{Feedback C $\rightarrow$ B\}} Using other agents' solutions and feedbacks as additional information, can you provide your answer to the math problem? The original math problem is \{Question\} Your final answer should be a single numerical number, in the form \textbackslash boxed\{answer\}, at the end of your response.  \\ 

\cmidrule{2-2}

    & \textbf{A}: Based on the correct reasoning and solutions provided by the other agents, the total amount of money Elise earned in the second year is \$1300. Therefore, the answer is \textbackslash boxed\{\green{1300}\}.    \\

\cmidrule{1-2}

\multirow{7.5}{*}{\textbf{Agent C}}

    & \textbf{Q}: Here are the feedbacks for your solution from the above two agents: One agent feedback: \texttt{\{Feedback A $\rightarrow$ C\}} One agent feedback: \texttt{\{Feedback B $\rightarrow$ C\}} Using other agents' solutions and feedbacks as additional information, can you provide your answer to the math problem? The original math problem is \{Question\} Your final answer should be a single numerical number, in the form \textbackslash boxed\{answer\}, at the end of your response.  \\ 

\cmidrule{2-2}

    & \textbf{A}: Based on the correct reasoning provided, the total amount of money Elise earned in the second year is \$1300. Therefore, the final answer is \textbackslash boxed\{\green{1300}\}.  \\

\bottomrule
\end{tabular}
\caption{\textbf{Stage 3: Revise.} Upon receiving peer reviews, agents revise their initial solutions. All agents arrive at the correct answer.} 
\label{tab:stage3}
\end{table*}

\newpage

\end{document}